\documentclass[11pt,a4paper]{article}
\usepackage[hyperref]{emnlp2020}
\usepackage{times}
\usepackage{latexsym}

\usepackage{microtype}

\usepackage{microtype}

\aclfinalcopy 
\def\aclpaperid{3032} 

\usepackage{soul}
\usepackage{url}
\usepackage{graphicx}
\usepackage{amsmath}
\usepackage{amsthm}
\usepackage{booktabs}
\usepackage{algorithm}
\usepackage{algorithmic}
\usepackage{xcolor}
\usepackage{xspace}

\newcommand{\ourdata}[1]{\textsc{OpenPI}}

\definecolor{Red}{rgb}{1,0,0}
\definecolor{Green}{rgb}{0,0.7,0}
\definecolor{Blue}{rgb}{0,0,1}
\definecolor{Red}{rgb}{0.6,0,0}
\definecolor{Orange}{rgb}{1,0.5,0}

\newcommand{\camera}[1]{#1}

\makeatletter
\newcommand*\bigcdot{\mathpalette\bigcdot@{.5}}
\newcommand*\bigcdot@[2]{\mathbin{\vcenter{\hbox{\scalebox{#2}{$\m@th#1\bullet$}}}}}
\makeatother

\newcommand{\bluebox}[1]{\colorbox{blue!10}{#1}}
\newcommand{\redbox}[1]{\colorbox{red!10}{#1}}

\usepackage{amsmath}  

\newcommand{\com}[1]{}

\newcommand{\eat}[1]{}
\mathchardef\mhyphen="2D

\newcommand{\blue}[1]{\textcolor{blue}{#1}}

\newcommand{\squishlist}{
  \begin{list}{$\bullet$}
    { \setlength{\itemsep}{0pt}      \setlength{\parsep}{3pt}
      \setlength{\topsep}{3pt}       \setlength{\partopsep}{0pt}
      \setlength{\leftmargin}{1.5em} \setlength{\labelwidth}{1em}
      \setlength{\labelsep}{0.5em} } }
\newcommand{\reallysquishlist}{
  \begin{list}{$\bullet$}
    { \setlength{\itemsep}{0pt}    \setlength{\parsep}{0pt}
      \setlength{\topsep}{0pt}     \setlength{\partopsep}{0pt}
      \setlength{\leftmargin}{0.2em} \setlength{\labelwidth}{0.2em}
      \setlength{\labelsep}{0.2em} } }
      
 \newcommand{\squishend}{
     \end{list} 
 }

\newcommand{\veryshortarrow}[1][4pt]{\mathrel{%
   \hbox{\rule[\dimexpr\fontdimen22\textfont2-.2pt\relax]{#1}{.4pt}}%
   \mkern-4mu\hbox{\usefont{U}{lasy}{m}{n}\symbol{41}}}}

\newcommand{\action}[1]{\ul{\textsl{#1}}}

\usepackage{soul}

\newcommand{\anycond}[1]{\textsl{#1}}

\newcommand{\precond}[1]{\anycond{#1}}

\newcommand{\attr}[1]{\emph{#1}}
\newcommand{\entity}[1]{\texttt{#1}}
\newcommand{\postcond}[1]{\anycond{#1}}
\newcommand{\yset}[1]{$\boldsymbol{y}$}

\newcommand{\condpair}[2]{\precond{#1} $\veryshortarrow$ \postcond{#2}}

\newcommand{\yipre}{$y_{i}^{\textit{pre}}$\xspace}
\newcommand{\yipost}{$y_{i}^{\textit{post}}$\xspace}

\definecolor{light-gray}{rgb}{0.95703125,0.953125,0.94921875}

\title{A Dataset for Tracking Entities in Open Domain Procedural Text}

\author{Niket Tandon{\normalfont \textsuperscript{1}}  

\quad Keisuke Sakaguchi{\normalfont \textsuperscript{1}} 

\quad  Bhavana Dalvi Mishra{\normalfont \textsuperscript{1}} 

\quad Dheeraj Rajagopal {\normalfont \textsuperscript{2}} \\ 

{\bf Peter Clark{\normalfont \textsuperscript{1}}} 

\quad {\bf Michal Guerquin{\normalfont \textsuperscript{1}}} 

\quad {\bf Kyle Richardson{\normalfont \textsuperscript{1}}} 

\quad {\bf Eduard Hovy{\normalfont \textsuperscript{2}}} 
\\
    
\textsuperscript{1}Allen Institute for Artificial Intelligence, Seattle, WA \\

{\tt \{nikett, keisukes, bhavanad, peterc, michalg, kyler\}@allenai.org} \\

\textsuperscript{2}Department of Computer Science, CMU\\ 

{\tt \{dheeraj, hovy\}@cs.cmu.edu}
}

\date{}
\begin{document}
\maketitle

\begin{abstract}
We present the first dataset for tracking state changes in procedural text from arbitrary domains by using an {\it unrestricted} (open) vocabulary.
For example, in a text describing fog removal using potatoes, a \entity{car window} may transition between being \attr{foggy, sticky, opaque}, and \attr{clear}.
Previous formulations of this task provide the text and entities involved, and ask how those entities change for just a small, pre-defined
set of attributes (e.g., location), limiting their fidelity. \camera{Our solution is a new task formulation where given just a procedural text as input, the task is to generate a set of state change tuples {\it (entity, attribute, before-state, after-state)} for each step, where the entity, attribute,
and state values must be predicted from an open vocabulary.} Using crowdsourcing, we create \ourdata~\footnote{\camera{Download \ourdata~ at \url{https://allenai.org/data/openpi}}}, a high-quality  \camera{(91.5\% coverage as judged by humans and completely vetted)}, and large-scale dataset comprising \camera{29,928} state changes over 4,050 sentences from 810 procedural real-world paragraphs from WikiHow.com. A current state-of-the-art generation model on this task \camera{achieves 16.1\% F1 based on BLEU metric}, leaving \camera{enough} room for novel model architectures.
\end{abstract}

\section{Introduction}
\label{sec:intro}

By one estimate, only about 12\% of what we understand from text is expressed explicitly  \cite{Graesser1981ProseCB}. This is especially apparent in text about actions where the effects of actions are left unstated. Humans fill that gap easily with their commonsense but machines need to model these effects in the form of state changes. For example, when 
\action{a potato is rubbed on a car window} (to defog it), then the unstated effects of this action are the following state changes: \anycond{windows becomes sticky, opaque, and the potato becomes dirty}, etc. These changes can be tracked across the paragraph. 
An exemplary use case of text with actions is procedural text (recipes, how-to guides, etc.) where modeling such state changes helps in various reasoning-based end tasks, e.g. automatic execution of biology experiments \cite{mysore2019materials}, cooking recipes \cite{Bollini2012cookingrobot} and everyday activities \cite{Wikihow-sigir2015}.

\begin{figure}[!t]{\includegraphics[width=1.0\columnwidth]{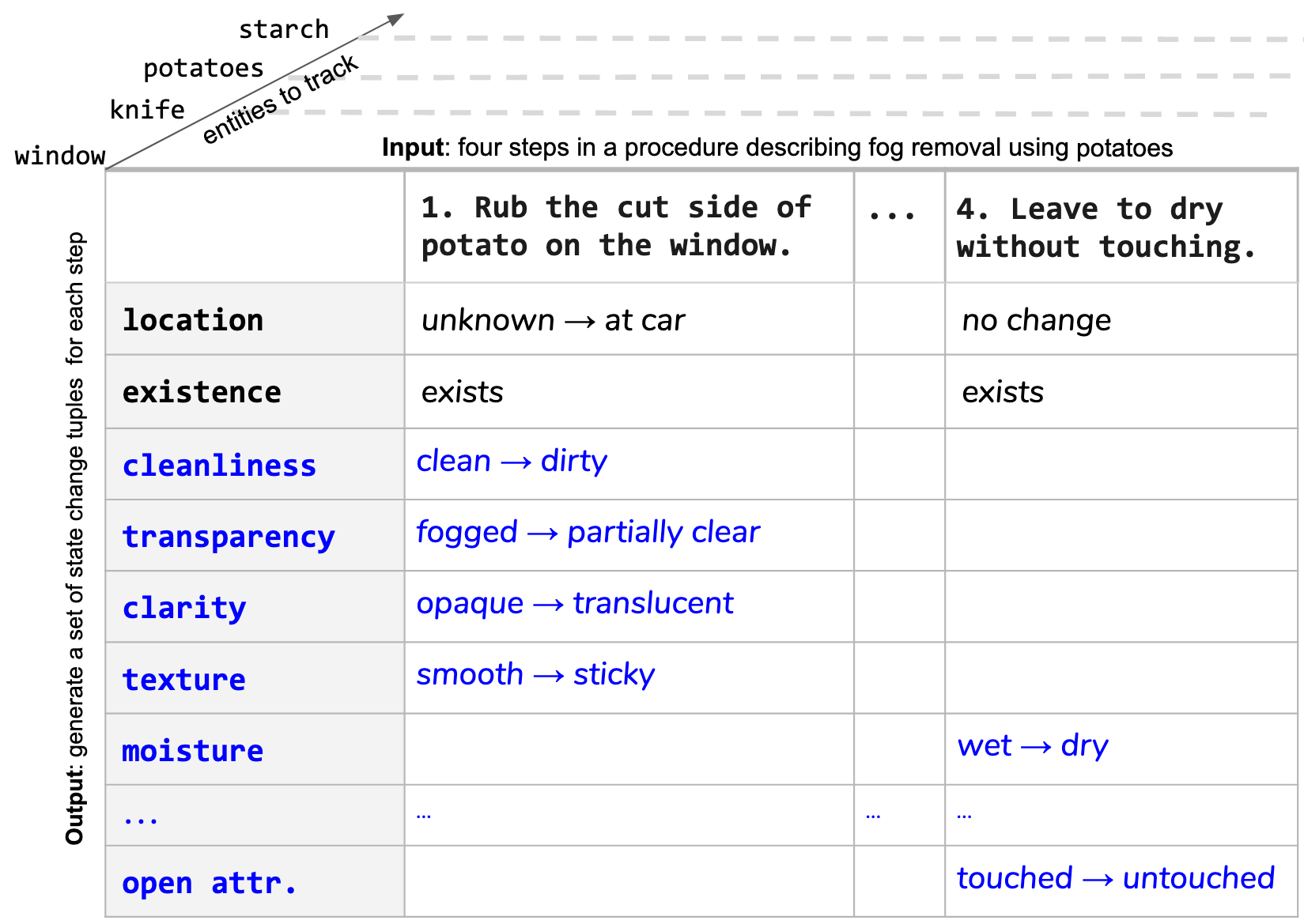}}
\caption{Previous formulations of the state tracking task are restricted to a small, fixed set of pre-defined state change types that limits its fidelity to model real-world procedures (they cannot cover the blue part in this procedure comprising four steps). Our solution is a new task formulation to track an unrestricted (open) set of state changes (additionally covering \blue{blue}). 
}
\label{fig:running-example}
\end{figure}


While there has been great progress in tracking entity states in scientific processes \cite{propara-naacl18}, tracking ingredients in cooking recipes \cite{npn}, and tracking the emotional reactions and motivations of characters in simple stories \cite{Rashkin2018psychology}, prior tasks are restricted to a fixed, small set of state change types thus covering only a small
fraction of the entire world state. Figure \ref{fig:running-example} illustrates this for a real-world procedure ``How to Keep Car Windows Fog Free Using a Potato''. Existing datasets such as ProPara \cite{propara-naacl18} only model the \attr{existence} and \attr{location} attributes, limiting the fidelity with which they model the world.
Specifically:
\squishlist
\item Attributes from domain-specific datasets such as ProPara \cite{propara-naacl18} and Recipes \cite{npn}  together, only cover $\sim$40\% of the state changes that people typically mention when describing state changes in real-world paragraphs from WikiHow (\S \ref{sec:challenges}).
\item The set of attributes that people naturally use to describe state changes is large,
  and hence hard to pre-enumerate ahead of time (especially when the target domain is unknown).
  Even a comprehensive list of popular attributes failed to cover 20\% of those used in practice (\S \ref{sec:dataset-stats}).
\item The dominant approach in existing datasets is to assume that changing entities are mentioned as spans in the procedural text.
  However, in unconstrained human descriptions of changes, $\sim$40\% of the referred-to entities were unmentioned in the text (e.g., the knife and cutting board in several cooking recipes) (\S \ref{sec:dataset_quantative_analysis}).
\squishend

  Addressing these limitations, our solution is a new task formulation to track an unrestricted (open) set of state changes:
  Rather than provide the text and entities, and ask how those entities change for a pre-defined set of attributes at each step,
  we instead provide just the input text, and ask for the set of state changes at each step, 
  each describing the before and after values of an attribute of an entity in the form (\anycond{$attribute$ of $entity$ was $value_{\text{before}}$ before and $value_{\text{after}}$ afterwards.}). Importantly, the vocabularies for attributes, entities, and values is open (not pre-defined).
%
%
%
%
%
%
%
%
%
%
%
\begin{table*}[!h]
\centering
\resizebox{1.0\textwidth}{!}
{
    \begin{tabular}{|p{5cm}|p{11cm}|}
    \hline
    
    \vspace{0pt} & \\
    \bf{Input $x$} & \textbf{Output \yset}\\
    \vspace{0pt} & \\
    \hline

    \action{Apply insecticide to peonies.} & \precond{the location of insecticide was in bottle before and on peonies afterwards.} \\
    &  \precond{the health of bugs were healthy before and dying afterwards.} \\
    
    & \\
    
    \action{Dip the peony flowers in water.} & \precond{the moisture of flowers was dry before and wet afterwards.} \\
     &  \precond{the cleanliness of peonies were dirty before and clean afterwards.} \\
    
    & \\
    \action{Stop ants from climb.. use trap} & \precond{the organization of trap was disassembled before and assembled after..} \\
    &  \precond{the well being of plants were troubled before and healthy afterwards.} \\

    \hline

    & \\

    \action{Combine apricots, .. in blender.} & \precond{the location of apricots was on counter before and in blender afterwards.} \\
    &  \precond{the state of ingredients were separate before and combined afterwards.} \\
    &  \precond{the weight of blender was light before and heavy afterwards.} \\

& \\

    \action{Add oil until dressing  thick.} & \precond{the state of ingredients were separate before and combined afterwards.} \\
    &  \precond{the location of oil was on counter before and in blender afterwards.} \\

& \\

    \action{Stir in the basil.} & \precond{the location of dressing was in blender before and on serving plate..} \\
    &  \precond{the location of basil was outside blender before and in blender afterwards.} \\
    &  \precond{the weight of blender was heavy before and light afterwards.} \\

    \vspace{0pt} & \\
    
    \hline
    \end{tabular}
}
\caption{Examples of the task based on our dataset. The input $x$ comprises a {\bf \action{query} $x_q$} and a context $x_c$ (past sentences before this step in the paragraph-- not shown due to limited space). The output is a set \yset~ of pre and postconditions. The paragraphs in this table are: above (how to clean oven) and below (cooking recipe). }
\label{tab:many-examples}
\end{table*}
Our contributions are:
\squishlist
\item[(i)] we introduce a novel task of tracking an unrestricted (open) set of state change types (\S \ref{sec:task}). 
\item[(ii)] we create a large-scale ($\sim$30K state changes), high-quality \camera{$\sim$ 91.5\% coverage and human vetted)} crowdsourced annotated dataset 
\ourdata~, from a general domain text serving as training dataset for this task (\S \ref{sec:dataset}).
\item[(iii)] we establish a strong generation baseline demonstrating the difficulty of this task  (\S \ref{sec:model}), and present an error analysis suggesting avenues for future research (\S \ref{sec:error_analysis}). 
\squishend

\section{Proposed Task: \ourdata}
\label{sec:task}


From a procedural paragraph with sentences (i.e., steps) ${step}_1 \dots {step}_K$, construct $K$ data points, one per step.

\paragraph{Input:} 
As input we are given a procedural text comprising current step ${step}_i$ as query and all past step as context ${step}_1 \cdots {step}_{i-1}$. We denote the input as $x=(x_q, x_c)$, where $x_q$ is the step for which we need the state changes (i.e. the query) and $x_c$ is the context.

Here, we use the common assumption \cite{propara-naacl18} that the steps in procedural text are ordered such that the context required for ${step}_i$ is mentioned in ${step}_1 \cdots {step}_{i-1}$.

\paragraph{Output:}
The output is a set of zero or more state changes \yset~$=\{y_i\}$. A state change $y_i$ is of the form: \bluebox{\anycond{$attr$ of $ent$ was $val_{pre}$ before and $val_{post}$ afterwards}}\\

Here, $attr$ is the attribute or state change type, and $ent$ is the changed entity. $val_{pre}$ is the precondition (i.e., the state value before), and $val_{post}$ is the postcondition (i.e., the state value afterwards). Pre/ postcondition  $adj\_or\_relp(y_i^{pre})$ can be an adjectival phrase or a relational phrase. In this task, $attr$, $ent$, $val_{pre}$ and $val_{post}$ are open form text i.e. they are not tied to any fixed, constrained vocabulary.

\paragraph{Example:} Consider the running example: $x$=(context: \anycond{The window of your car is foggy}, query: \action{Rub half potato on the window}). Then, $\{y\} = \{$ \anycond{transparency of window was fogged before and partially clear afterwards}, \anycond{stickiness of window was smooth before and sticky afterwards} $\}$. In $y_1$, $attr$ = \attr{transparency}, $ent$ = \entity{window}, $val_{pre}$ = \anycond{fogged} and  $val_{post}$ = \anycond{partially clear}





\subsection{Unique Challenges} 
\label{sec:challenges}
\ourdata~ has two unique challenges that are not found in any existing state change dataset. 

\squishlist
\item \textbf{Variable size, low-specificity output}: \cite{imagespecificity} introduce the notion of image specificity which measures the amount of variance in multiple viable descriptions of the same image (typically, each image has exactly $K$ descriptions from $K$ annotators). Low specificity implies very different descriptions that are not mere re-phrasings. In \ourdata, the output \yset~ has low-specificity (low specificity is also called high complexity output). To achieve low specificity outputs, existing methods learn to generate diverse responses by sampling different keywords and using a reinforcement learning approach for training \cite{gao2019generatingdiverse} or use a diverse beam search \cite{kalyan2016diversekbeam} based approach on a typical encoder to decode diverse outputs. However, they all assume that the output set size is fixed to $K$ (typically each sample is annotated by exactly $K$ annotators). In our case, however, the number of items in \yset~ is variable, making these existing solutions inapplicable.

\item \textbf{Open vocabulary}: In \ourdata, $attr$, $ent$, $val_{pre}$ and $val_{post}$ are not restricted to any fixed, small  vocabulary. Previous task formulations such as  \cite{npn,propara-naacl18}, made the assumption that $ent$ is given, $attr$ is from a vocabulary of less than 10 classes, and $val_{pre}$ or $val_{post}$ are either from a small external vocabulary or a span in $x$\footnote{We matched an exhaustive list of synonyms of existing attributes from existing datasets ProPara and Recipes: \attr{existence, location, temperature, composition, cleanliness, cookedness, shape, rotation, accessibility} and found that only $\sim$40\% of the attributes in \ourdata~ are covered by these (however,  these datasets cannot cover the open vocabulary of entities and attribute values)}. In contrast, in \ourdata~, the entities may not be present in the sentence or even the context, and the state change types and values can come from a rather open vocabulary. This openness brings a variety of challenges: (i) presupposed entities: these are entities that are not present in $x$ and perceived through background knowledge, (ii) zero shot learning: during inference on a previously unseen domain, there are previously unseen attributes, entities, and state change types. This makes the problem very challenging and places this task in a novel setting (see \S \ref{tab:subsec_positioning}) 





\squishend


\begin{table}[]
\centering
\begin{tabular}{llll}
\toprule
Task & Vocab. & Specificity & Output \\ 
     &       &             & size \\ 
\midrule
Story CSK & open & high & fixed \\
ProPara & closed & high & fixed \\
Recipes Task & closed & low & fixed \\
ALFRED & closed & high & fixed \\
VirtualHome & closed & high & fixed \\
OpenPI & open & low & variable \\ \bottomrule
\end{tabular}
\caption{Comparison of our dataset to existing datasets}
\label{tab:related_work}
\end{table}

\label{fig:related-work}

\section{Related Work}
\label{sec:related-work}


\textbf{Tracking state changes:}
Procedural text understanding addresses the task of tracking entity states throughout the text \cite{npn,Henaff2016TrackingTW}. This ability is an important part of text understanding. While syntactic parsing methods such as AMR (abstract meaning representation) \cite{AMR} represent ``who did what to whom'' by uncovering stated facts, tracking entity states uncovers unstated facts such as how ingredients change during a recipe. 

\textbf{Datasets with closed state changes:}
The bAbI dataset~\cite{weston2015towards} includes questions about objects moved throughout a paragraph, using machine-generated language over a deterministic domain with a small lexicon. The SCoNE dataset~\cite{long2016simpler} contains paragraphs describing a changing world state in three synthetic,  deterministic domains.  However, approaches developed using synthetic data often fail to handle the inherent complexity in language when applied to organic, real-world data~\cite{hermann2015teaching,winograd72}. The ProPara dataset \cite{propara-naacl18} contains three state changes \attr{(create, destroy, move)} for natural text describing scientific procedures. Other domain specific datasets include recipe domain \cite{npn}, and biology experiments \cite{mysore2019materials}. These datasets contain a small, closed set of state change types that are relevant to a specific domain. Our dataset is general domain, and to accommodate this generality we have an open vocabulary of state changes. 

\textbf{Datasets with open state changes:} \cite{Isola2015DiscoveringTransformationVision} propose manually defined antonymous adjective pairs (big, small) to define transformations in images, and this was an inspiration for us to use adjectives as open state changes in \ourdata. Knowledge bases such as ConceptNet  \cite{Speer2013ConceptNet5A} and ATOMIC \cite{Sap2019ATOMICAA} contain (open) pre-conditions and post-conditions but they are agnostic to context. Context plays a role when dealing with a large number of state changes types e.g., if ``a stone hits a glass'' then the glass would break but this is not the case if ``a soft toy or a sound wave hits the glass''. Our dataset contains context information, an important training signal for neural models.

Current knowledge bases (such as ATOMIC) contain social rather than physical effects. 
As a result, generation models trained on these knowledge bases incorrectly force the effects to be social. For example, COMET \cite{Bosselut2019COMETCT}, trained on ATOMIC data, when applied on ``Cans are tied together and transported to a recycling center'', incorrectly predicts\footnote{Manually inspecting the 45 predictions made by COMET on this sentence, we found only one partially correct prediction that the human has to get to the recycle center before.} \condpair{person goes to recycle center, Person needs to be arrested}{Person is arrested, gets dirty}.

\subsection{Positioning \ourdata}
\label{tab:subsec_positioning}
Figure \ref{fig:related-work} projects existing tasks and models along two different dimensions (open vocabulary, and variable-size low-specificity). We find that models bottom-left quadrant represents majority of the existing work on state changes such as ProPara \cite{propara-naacl18} and bAbI \cite{Babi-tasks}) in NLP community, and ALFRED \cite{Shridhar2019ALFRED} and VirtualHome \cite{VirtualHomeSH} in Computer Vision. Correspondingly many models exist in that space (\cite{propara-emnlp18}, \cite{npn}, \cite{Henaff2016TrackingTW}). Very few models exist that can predict either open vocab \cite{Rashkin2018psychology}, or variable size output \cite{npn}. However, no existing task has \ul{both} open vocabulary and variable-size low specificity-- placing \ourdata~ in a novel space. 

\section{Dataset}
\label{sec:dataset}


\subsection{Data Collection}
We set up a crowdsourcing task on Amazon Mechanical Turk where the annotators author the \yset{}$=\{y_i\}$ for every sentence of a \url{wikihow.com} article, filling in a sentence template for each $y_i$ as a guide.
WikiHow contains a wide variety of goals (e.g., \textit{how to wash dishes}) broken down into steps with detailed descriptions and pictorial illustrations,  spanning across 19 categories. We selected a diverse subset of six popular categories and focus on action-oriented articles\footnote{We exclude WikiHow articles with steps containing stative verbs such as \textit{know, see, want}, etc., and remove articles with too few (less than 4) or too many steps (7 or more). The selected categories are in Table \ref{tab:dataset-stats}.}.

For a given WikiHow article, annotators were asked to describe up to six state changes for each step ($0\leq|\text{\yset~}|\leq6$), and were paid fairly\footnote{We set the reward to be \$0.07 for each of the first three state changes, and \$0.14 for each of the additional three state changes in order to encourage workers to write as many state changes as possible. All annotators met the following prerequisites as a minimum qualification: (1) 5K previous HITs approvals, (2) 99\% or higher approval rate, (3) location is US, UK, CA, AU, or NZ.}. Each state change description consists of precondition (\yipre), postcondition (\yipost), and the (physical) attribute. \camera{Restricting the annotators to a template for state change described in \S\ref{sec:task}, yields much better quality than free-form. This was a pragmatic choice, to encourage Turkers to give a complete description but not add extra noise. In an earlier pilot, we tried upto 10 changes but Turkers found the task too difficult and complained. Six empirically resulted in the best level of agreement and completeness among annotations, while also retaining diversity.}

The annotators were encouraged (but not required) to pick from a pre-defined vocabulary of 51 WordNet derived attributes.

\begin{figure}[!h]
\begin{center}
\centerline{\includegraphics[width=1.05\columnwidth]{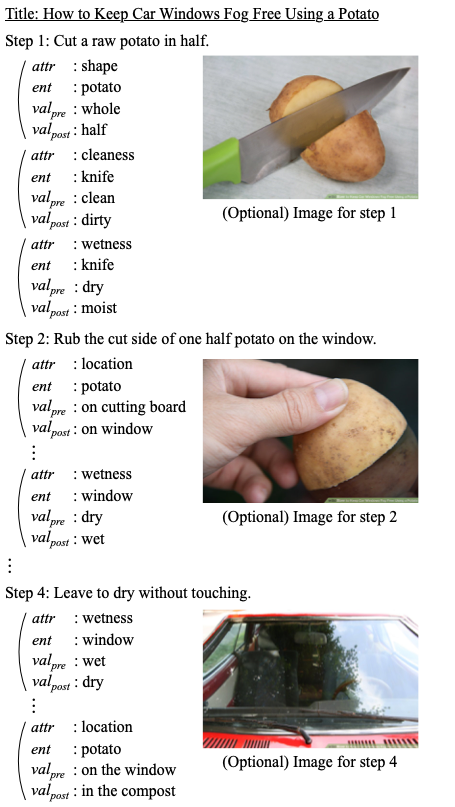}}
\caption{Data collection procedure: Crowdworkers are shown the article title, step descriptions and optionally the corresponding image, and asked to write up to six state changes (\yipre, \yipost, \attr{attr}) per step. See the appendix for a sample of the annotation task.}
\label{fig:hit}
\end{center}
\end{figure}

We performed two sets of annotations for every article, one where the annotators see the pictorial illustration of a step and one without. Visuals helped the annotators to provide more state changes (e.g., the color of cut potato turns gray). In total, one article is annotated four times (two turkers each for with and without images)-- making the cost of annotation \$3.6 in average per article. See Figure~\ref{fig:hit} for an example of the annotation procedure.

\camera{After collecting the data, we cleaned up the state changes by asking three crowd workers if each state change is valid or not with the same annotation setting as data collection (e.g., with or without visual illustration). We discarded state changes that did not get the agreement by the majority (2 or more workers). With this cleaning step, the total number of state changes changed from 33,065 to 29,928.}

\camera{The small number of errors encountered during vetting fell into five categories}:
\squishlist
\item ($\sim$45\% of the errors) Obscure attributes/ values, e.g., \anycond{state of clubhouse was spoken of before}.
\item ($\sim$20\%) State change of future steps, e.g.,
  \action{Prepare the pot}  $\rightarrow$ \anycond{location of veggies in pot}
\item ($\sim$15\%) Mismatch of attribute and value:
  \anycond{shape of lemon was solid}
\item ($\sim$10\%) State change of the reader, not the actor:
\anycond{knowledge of you becomes aware} 
\item ($\sim$10\%) Factual errors: annotated change does not occur or tautologously refers to the action.
\squishend

\subsection{Dataset statistics} \label{sec:dataset-stats}
The resulting \ourdata~ dataset comprises \camera{29,928} state changes over 4,050 sentences from 810 WikiHow articles. 
Of these, 
\camera{15,445 (4.3 per step)} state changes were obtained from the \textit{with images} setting and \camera{14,483 (3.8 per step)} from \textit{without images}, indicating that the additional visual modality helped workers to come up with more state changes (e.g., the color of cut potato turns gray). 
These WikiHow articles were from six categories, see Table \ref{tab:dataset-stats}. The number of state changes in a category depends on the density of entities and their changes e.g., cooking related articles include multiple ingredients and instruments that undergo state changes. 


\begin{table}[h]
    \setlength{\tabcolsep}{4pt}	
\begin{tabular}{l|rrrr}
\hline
WikiHow cat. & \multicolumn{1}{l}{\# para} & \multicolumn{1}{c}{$| \text{\yset~}|$} & \multicolumn{1}{l}{w/ img} & \multicolumn{1}{l}{w/o}  \\ \hline
 Food \& Entertain &     197      &  9942 &   5399   &  4543\\
    Home \& Garden    &     199      &  6961 &   3758   &  3203\\
   Hobbies \& Craft  &     193      &  4766 &   2375   &  2391\\
   Sports \& Fitness  &      95      &  3361 &   1662   &  1699\\
 Cars \& Vehicles &      43      &  1656 &   818    &  838\\
         Health        &      77      &  3036 &   1433   &  1603\\\hline
          All          &     858      & 29928 &  15445   & 14483\\\hline
\end{tabular}
\caption{\camera{Basic statistics of the \ourdata{} dataset: the articles' WikiHow category, the number of WikiHow articles (i.e., paragraphs) in each category and number of state changes $|\text{\yset~}|$ (total), and data collected using with, and without image setting).}}
\label{tab:dataset-stats}
\end{table}

Two thirds of the state changes are adjective phrases \camera{(avg. length 1.07 words)} and the remaining one third are relational phrases \camera{(avg. length 2.36 words)}. Attributes, entities, adjective phrases, relational phrases all follow a power-law like  distribution. The most popular adjectives were \{\attr{dry, empty, clean, wet, dirty, full, heavier, lighter, hot, whole, cool, cold}\}, and the most popular relational phrases were location-indicating prepositions. About 20\% of the attributes are present in 80\% of the data. The long tail of the remaining 80\% attributes \ul{indicates why} open attributes are important. As similar attributes can be expressed differently in text e.g., \attr{wetness} and \attr{moisture}, we analyzed a few data points to observe a large agreement between annotators in choosing attributes (the average size of attribute clusters was only 1.2).

\begin{figure*}[!ht]{\includegraphics[width=1.01\textwidth]{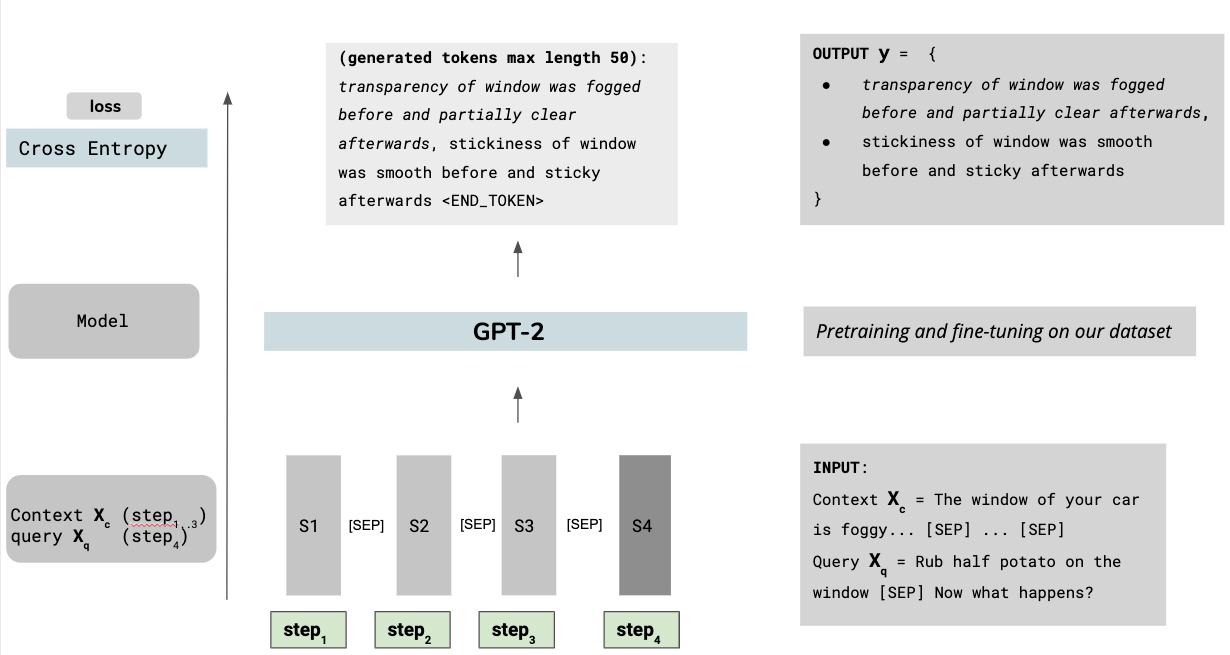}}
\caption{Our GPT-2 based model for \ourdata.} 
\label{fig:model}
\end{figure*}


We split the data into training, development, and test sets.
To evaluate model generalization, all the annotated articles in the Health category are marked as out-of-domain
and placed (only) in the test set.
All the remaining annotated articles are randomly assigned to one of the splits. The resulting training set \camera{has 
23,869 state changes (3,216 instances because one instance comprises $|\text{\yset~}|$ state changes), dev set has 
1,811 (274 instances), and test set has 
4,248 
(160 instances in domain, and 394 instances out-of-domain ``Health'')}. 

\subsection{Dataset quality}
\label{sec:dataset_quality}

\camera{We measure the quality (coverage) of the dataset by asking a human judge whether there is any new state change they can add.} \camera{The judge added only 8.5\% new state changes. This suggests that \ourdata{} has a high coverage of $\sim$91.5\%, and a very high precision because of vetting.} 

These additions fell into four categories:
\squishlist
\item ($\sim$40\% of additions) Indirect effect was missed, e.g.,
 \action{Place in freezer} $\rightarrow$ (existing) \anycond{food cooler}, (added) \anycond{food container cooler}
\item ($\sim$35\%) Extra dimension of change (attribute) missed, e.g., (added) Change in texture, organization, open/closed state.
\item ($\sim$20\%) Addition is a rewording hence not helpful e.g.,
\anycond{cleanliness of windshield}, (added) \anycond{clarity of windshield}
\item ($\sim$5\%) Addition is incorrect/obscure.
\squishend

\subsection{Quantifying the reasoning challenges} 
\label{sec:dataset_quantative_analysis}



\textbf{Presupposed entities:}
About 61\% of the entities in our development set are mentioned as spans in the context and paragraph, while the remaining 40\% are unmentioned entities. About 35\% of the unmentioned entities were derivatives of mentioned entities, i.e. synonym, hypernym-hyponym, or part-whole. The remaining 65\% were presupposed (assumed) entities, e.g., containers of mentioned entities, surfaces, cooking instruments. 

\textbf{Open attributes:} 78.9\% of the examples contain the 51 predefined attributes that the annotators were supplied. The remaining examples contain 577 Turk authored open attributes and many of these are difficult to anticipate, e.g., cookedness, tightness, saltiness. This makes up a long tail distribution of an open vocabulary of attributes. 

\textbf{Zero-shot learning:} 
The test-set contains: 1) paragraphs from five categories covered in the training set, 2) paragraphs from  \texttt{Health} category for which there is no training data, to test zero-shot learning. \texttt{Health} test-subset is particularly challenging with 55\% unmentioned entities (40\% otherwise) and 33\% unseen attributes (18\% otherwise).

\textbf{Variable size, low specificity output:} A system needs to decide relevant entities and attributes would be relevant and generate possibly varying number of state changes for different steps. The dev set has on average seven state changes per step, and 3\% of the steps have no state change. 



\section{Model}
\label{sec:model}

\ourdata{} dataset poses unique challenges including presupposed entities, open attributes, zero-shot learning and variable-size, low specificity output (see Section \ref{sec:dataset_quantative_analysis}). These challenges make it difficult to apply existing entity tracking methods like ProStruct \cite{propara-emnlp18}, EntNet \cite{Henaff2016TrackingTW}, NPN \cite{npn} without making significant changes to either the model or the task. E.g., the commonsense constraints in ProStruct do not scale with a large number of attributes, and EntNet is not suitable for a set output. 

\ourdata{} is well-suited for a generation model because the output \anycond{$attr$ of $ent$ was $val_{pre}$ before and $val_{post}$ afterwards} must be predicted using an open vocabulary. Therefore, as our baseline, we use the state-of-the-art pre-trained language model, GPT-2 \cite{gpt2}, and fine-tune it for \ourdata{} task. The model takes as input a special [SEP] token separated $x_c$ and $x_q$. The output is expected to be a set \yset~ of variable size. As noted in \S \ref{sec:challenges}, existing methods do not produce a variable size, low-specificity output. Instead we train the model to generate a long sequence of comma separated $y_i$. If there are no changes i.e., $|\text{\yset~}|=0$, then we set \yset~ $= \{$\anycond{there will be no change}$\}$.


Figure \ref{fig:model} shows the model architecture. During decoding, we sample $y_i$ as a sequence of output tokens generated by the model. The generation output accounts for all aspects of the state change - the attribute, entity, and before, after values.
\section{Experiments}


\subsection{Metrics}
\label{sec:metrics}
To measure the performance on \ourdata, we compare the predicted set \yset~ and gold set \yset{}*, for every point $x$. 
Precision for a data point $x$ is computed based on the best matching gold state change for each predicted state change i.e., $P(x) = \frac{1}{2}~ \sum_{y \in \text{\yset~}} max_{y*}~ O(y*^{pre}, y^{pre}) + O(y*^{post}, y^{post}) $. Similarly, recall is based on the best matching predicted state change for each gold state change i.e. $R(x) = \frac{1}{2}~ \sum_{y* \in \text{\yset~*}} max_{y}~ O(y*^{pre}, y^{pre}) + O(y*^{post}, y^{post}) $. The string overlap function $O(.)$ can use any of the standard generation metrics: exact match, BLEU, METEOR or ROUGE\footnote{ \url{github.com/allenai/abductive-commonsense-reasoning}}. We report micro-averaged precision, recall, F1 scores for different choices of $O(.)$.

\camera{We remove template words before string comparison 
to avoid inflating scores for template words. We did not perform facet-based evaluation of the templated output for two reasons. Firstly, while it might seem when computing overlaps of gold and predicted state changes as two long strings, BLEU or ROUGE may accidentally see an overlap when there was none. That is unlikely in practice because the entities, attributes, and values are quite distinct and scoring accidental overlaps is uncommon. Secondly, our evaluation metric (F1, precision, recall) matches a list of predictions against a list of gold references. It is unclear how to compute F1 over individual facets that requires the best match based on all facets as tuple.}

\camera{We also found that when manually evaluating on $\sim$200 dev datapoints, the score was systematically a few ($\sim$10\%) points higher than BLEU, while the trends and model rankings remained the same, indicating robustness of the automatic metric.}

\camera{Therefore, the proposed metric aligns with human evaluation, and is able to use existing generation metrics thereby simplifying evaluation, allowing easier reproducibility.}


\subsection{Evaluation}
\camera{We evaluate state of the art generation model GPT-2 on \ourdata{} dataset.  As mentioned in Section \ref{sec:dataset-stats}, \ourdata{} consists of two kinds of annotations: with-images (turkers were shown images along with text for each step of the procedure) and without-image (turkers looked at only text to predict state changes). GPT-2 gets to see only text as input but the state changes it has to predict are different depending on the setting. Table \ref{tab:f1-score} reports P, R and F1 when GPT-2 model is tested on different subsets.}

The GPT-2 model struggles to predict the right set of state changes indicating that the task is hard. Challenges include lexical variation on entities (in context vs. in gold), unseen categories, limited context for the initial sentences in the paragraph an so on. Detailed error analysis is presented in \S \ref{sec:error_analysis}.

\begin{table}[!h]
    \centering
    \resizebox{1.05\columnwidth}{!}{
    \begin{tabular}{l|ccc}
           & \multicolumn{3}{c}{F1 based on } \\ 
        & Exact & BLEU  &  ROUGE\\\midrule
with-image & 5.1 &	14.3	& 29.1\\ 
without-image & 3.6	& 13.4	& 28.2 \\\midrule
Entire dataset & 4.3	& 16.1 &	32.4  \\
    \end{tabular}
    }
    \caption{\camera{GPT-2 on OpenPI, and its sub-categories.}}
    \label{tab:f1-score}
\end{table}



\begin{table}[!h]
    \centering

    \begin{tabular}{l|ccc}
        Models   & \multicolumn{3}{c}{BLEU scores} \\ 
        & P & R & F1 \\\midrule
seen category	& 25.1	& 18.4	& 17.1 \\
unseen categories &	24.4 &	17.4 &	15.7 \\
    \end{tabular}
    \caption{\camera{GPT-2 on topics seen,  unseen during training.}}
    \label{tab:seen_unseen}
\end{table}

\ourdata{} testset comprises of both unseen and seen categories, and we report BLEU  separately on these subsets. Results from table \ref{tab:seen_unseen} 
presents an encouraging result that GPT-2 generalizes to unseen topics even though the scores on seen categories is understandably a little higher (F1 of 17.1 for seen category vs  15.7 for unseen categories).

\subsection{Error analysis} \label{sec:error_analysis}
To better understand model shortcomings, the error types in dev predictions are illustrated (Table \ref{tab:error_types}).
\begin{table}[!h]
    \centering

    \begin{tabular}{l|cc}
      Error type 
                 & freq & \% \\\midrule
Wrong attribute & 826 & 51\\
Wrong entity  &	964 & 59\\
Wrong adjective 	& 989 & 41\\
Wrong relation phrase	& 456 & 17\\ \midrule
Any of the above 	& 1,622 & 100\\

    \end{tabular}
    \caption{\camera{Error types in 1,811 dev predictions. One state change prediction can have multiple error types.}}
    \label{tab:error_types}
\end{table}



\begin{enumerate}
\item Wrong attribute ($attr(y_i)$): In 51\% state changes produced by the GPT-2 model, predicted attribute is incorrect. Often ($\sim$20\% of cases) predicted attribute is \attr{state}, i.e. the model couldn't name the attribute.
	
\noindent\fbox{
    \small
    \parbox{0.44\textwidth}
    {
      \reallysquishlist
      
        \item[Gold:] \bluebox{wetness of potatoes was wet before, and dry after}\item[Pred:] \redbox{state of potatoes was wet before, and dry after}
      \squishend
    }}
\item Wrong entity ($ent(y_i^{pre})$): The model predicted incorrect entity 59\% of the times. 
For 32\% of the entity errors, the gold entity was unmentioned in the input text.


\squishlist
\item[(i)] \camera{Entities present as span (68\%): Typically, a related but not same entity is predicted:}

    \noindent\fbox{
    \small
    \parbox{0.42\textwidth}
    {
      \reallysquishlist
        \item[G:] \bluebox{..furniture was worn out before, and renewed after}\item[P:] \redbox{..chairs was dirty before, and clean after}
      \squishend
    }}
    
\item[(ii)] Derivable entities: (3\%)
These entities are typically a lexical variation of the entities in the paragraph. 
E.g., \action{spray paint silk floral arrangement to change color or freshen its hue}, the model predicted
    \noindent\fbox{
    \small
    \parbox{0.32\textwidth}
    {
    G: \bluebox{..plant was dry before, and wet after}\\
      P: \redbox{..cloth was dry before, and wet after}
    }}
    

The following example also mentions a derivable entity and both gold and prediction are imply the same but it is difficult to automatically check that. E.g., \action{Keep the craft steady as others board.}

        \noindent\fbox{
    \small
    \parbox{0.39\textwidth}
    {
       G: \bluebox{stability of boat was rocking ... steadied after}\\
       P: \redbox{stability of craft was wobbling ... steady after}
    }}

    
    
\item[(iii)] Unmentioned entities: (29\%). These types of errors are very difficult to overcome because the entities are typically not mentioned at all in the generated output. For instance in the following, loser and rider both refer to the same person in the text,
    \noindent\fbox{
    \small
    \parbox{0.35\textwidth}
    {
      G: \bluebox{..loser was alive before, and dead after}\\
        P: \redbox{..rider was alive before, and killed after}
    }}
    
In about 20\% of such erroneous predictions, the model  predicted the $adj(y_i^{pre})$ correctly. This may be because attribute is a good indicator of the adjectives. 

\squishend

\item Wrong $adj(y_i^{pre})$ : (41\%) 
The model predicts incorrect adjectives, such that in some cases the erroneous adjectives might not apply to the given entity, or the adjective values are swapped between pre and post condition. An example is shown below:
    \noindent\fbox{
    \small
    \parbox{0.4\textwidth}
    {
      \reallysquishlist
        \item[G] \bluebox{..curtains was white before, and painted after}\item[P:] \redbox{..double curtains was colorless ...  colorful after}
      \squishend
    }}

\item Wrong $relp(y_i^{pre})$ (17\%): We find that relational phrases are very hard for the model currently. 184 out of 210 relational state changes predicted by the model have incorrect relational phrase. We believe that this poses a challenging research problem for future models. 
    \noindent\fbox{
    \small
    \parbox{0.415\textwidth}
    {
      \reallysquishlist
        \item[G:] \bluebox{knowledge of animals was absent ... present after}\item[P:] \redbox{details afterwards was ignored ... discussed after}
      \squishend
    }}
    


\item Length of the context plays an important role. Without any context (e.g., for the first step), the model gets a low accuracy of 8.3\%.


\end{enumerate}

\section{Conclusion}
We presented the first dataset to track entities in open domain procedural text. To this end, we crowdsourced a large, high-quality dataset with examples for this task. We also established a strong generation baseline highlighting the difficulty of this task. As future work, we will explore more sophisticated models that can address the highlighted shortcomings of the current model. An exciting direction is to leverage visuals of each step to deal with unmentioned entities and indirect effects.

\bibliography{the_bib_file}

\begin{thebibliography}{25}
\expandafter\ifx\csname natexlab\endcsname\relax\def\natexlab#1{#1}\fi

\bibitem[{Banarescu et~al.(2013)Banarescu, Bonial, Cai, Georgescu, Griffitt,
  Hermjakob, Knight, Koehn, Palmer, and Schneider}]{AMR}
Laura Banarescu, Claire Bonial, Shu Cai, Madalina Georgescu, Kira Griffitt, Ulf
  Hermjakob, Kevin Knight, Philipp Koehn, Martha Palmer, and Nathan Schneider.
  2013.
\newblock Abstract meaning representation for sembanking.
\newblock In \emph{LAW@ACL}.

\bibitem[{Bollini et~al.(2012)Bollini, Tellex, Thompson, Roy, and
  Rus}]{Bollini2012cookingrobot}
Mario Bollini, Stefanie Tellex, Tyler Thompson, Nicholas Roy, and Daniela Rus.
  2012.
\newblock Interpreting and executing recipes with a cooking robot.
\newblock In \emph{ISER}.

\bibitem[{Bosselut et~al.(2018)Bosselut, Levy, Holtzman, Ennis, Fox, and
  Choi}]{npn}
Antoine Bosselut, Omer Levy, Ari Holtzman, Corin Ennis, Dieter Fox, and Yejin
  Choi. 2018.
\newblock Simulating action dynamics with neural process networks.
\newblock \emph{ICLR}.

\bibitem[{Bosselut et~al.(2019)Bosselut, Rashkin, Sap, Malaviya, Çelikyilmaz,
  and Choi}]{Bosselut2019COMETCT}
Antoine Bosselut, Hannah Rashkin, Maarten Sap, Chaitanya Malaviya, Asli
  Çelikyilmaz, and Yejin Choi. 2019.
\newblock Comet: Commonsense transformers for automatic knowledge graph
  construction.
\newblock In \emph{ACL}.

\bibitem[{Dalvi et~al.(2018)Dalvi, Huang, Tandon, Yih, and
  Clark}]{propara-naacl18}
Bhavana Dalvi, Lifu Huang, Niket Tandon, Wen-tau Yih, and Peter Clark. 2018.
\newblock Tracking state changes in procedural text: A challenge dataset and
  models for process comprehension.
\newblock \emph{NAACL}.

\bibitem[{Gao et~al.(2019)Gao, Bi, Liu, Li, and Shi}]{gao2019generatingdiverse}
Jun Gao, Wei Bi, Xiaojiang Liu, Junhui Li, and Shuming Shi. 2019.
\newblock Generating multiple diverse responses for short-text conversation.
\newblock In \emph{AAAI}.

\bibitem[{Graesser(1981)}]{Graesser1981ProseCB}
Arthur~C. Graesser. 1981.
\newblock Prose comprehension beyond the word.
\newblock In \emph{Springer}.

\bibitem[{Henaff et~al.(2017)Henaff, Weston, Szlam, Bordes, and
  LeCun}]{Henaff2016TrackingTW}
Mikael Henaff, Jason Weston, Arthur Szlam, Antoine Bordes, and Yann LeCun.
  2017.
\newblock Tracking the world state with recurrent entity networks.
\newblock In \emph{ICLR}.

\bibitem[{Hermann et~al.(2015)Hermann, Kocisky, Grefenstette, Espeholt, Kay,
  Suleyman, and Blunsom}]{hermann2015teaching}
Karl~Moritz Hermann, Tomas Kocisky, Edward Grefenstette, Lasse Espeholt, Will
  Kay, Mustafa Suleyman, and Phil Blunsom. 2015.
\newblock Teaching machines to read and comprehend.
\newblock In \emph{Advances in Neural Information Processing Systems}, pages
  1693--1701.

\bibitem[{Isola et~al.(2015)Isola, Lim, and
  Adelson}]{Isola2015DiscoveringTransformationVision}
Phillip Isola, Joseph~J. Lim, and Edward~H. Adelson. 2015.
\newblock Discovering states and transformations in image collections.
\newblock \emph{CVPR}.

\bibitem[{Jas and Parikh(2015)}]{imagespecificity}
Mainak Jas and Devi Parikh. 2015.
\newblock Image specificity.
\newblock \emph{2015 IEEE Conference on Computer Vision and Pattern Recognition
  (CVPR)}, pages 2727--2736.

\bibitem[{Long et~al.(2016)Long, Pasupat, and Liang}]{long2016simpler}
Reginald Long, Panupong Pasupat, and Percy Liang. 2016.
\newblock Simpler context-dependent logical forms via model projections.
\newblock In \emph{ACL}.

\bibitem[{Mysore et~al.(2019)Mysore, Jensen, Kim, Huang, Chang, Strubell,
  Flanigan, McCallum, and Olivetti}]{mysore2019materials}
Sheshera Mysore, Zach Jensen, Edward Kim, Kevin Huang, Haw-Shiuan Chang, Emma
  Strubell, Jeffrey Flanigan, Andrew McCallum, and Elsa Olivetti. 2019.
\newblock The materials science procedural text corpus: Annotating materials
  synthesis procedures with shallow semantic structures.
\newblock \emph{arXiv preprint arXiv:1905.06939}.

\bibitem[{Puig et~al.(2018)Puig, Ra, Boben, Li, Wang, Fidler, and
  Torralba}]{VirtualHomeSH}
Xavier Puig, Kevin Ra, Marko Boben, Jiaman Li, Tingwu Wang, Sanja Fidler, and
  Antonio Torralba. 2018.
\newblock Virtualhome: Simulating household activities via programs.
\newblock \emph{2018 IEEE/CVF Conference on Computer Vision and Pattern
  Recognition}, pages 8494--8502.

\bibitem[{Radford et~al.(2019)Radford, Wu, Child, Luan, Amodei, and
  Sutskever}]{gpt2}
Alec Radford, Jeffrey Wu, Rewon Child, David Luan, Dario Amodei, and Ilya
  Sutskever. 2019.
\newblock Language models are unsupervised multitask learners.
\newblock \emph{OpenAI Blog}, 1(8):9.

\bibitem[{Rashkin et~al.(2018)Rashkin, Bosselut, Sap, Knight, and
  Choi}]{Rashkin2018psychology}
Hannah Rashkin, Antoine Bosselut, Maarten Sap, Kevin Knight, and Yejin Choi.
  2018.
\newblock Modeling naive psychology of characters in simple commonsense
  stories.
\newblock In \emph{ACL}.

\bibitem[{Sap et~al.(2019)Sap, Bras, Allaway, Bhagavatula, Lourie, Rashkin,
  Roof, Smith, and Choi}]{Sap2019ATOMICAA}
Maarten Sap, Ronan~Le Bras, Emily Allaway, Chandra Bhagavatula, Nicholas
  Lourie, Hannah Rashkin, Brendan Roof, Noah~A. Smith, and Yejin Choi. 2019.
\newblock Atomic: An atlas of machine commonsense for if-then reasoning.
\newblock In \emph{AAAI}.

\bibitem[{Shridhar et~al.(2019)Shridhar, Thomason, Gordon, Bisk, Han, Mottaghi,
  Zettlemoyer, and Fox}]{Shridhar2019ALFRED}
Mohit Shridhar, Jesse Thomason, Daniel Gordon, Yonatan Bisk, Winson Han,
  Roozbeh Mottaghi, Luke Zettlemoyer, and Dieter Fox. 2019.
\newblock Alfred: A benchmark for interpreting grounded instructions for
  everyday tasks.
\newblock \emph{ArXiv}, abs/1912.01734.

\bibitem[{Speer and Havasi(2013)}]{Speer2013ConceptNet5A}
Robyn Speer and Catherine Havasi. 2013.
\newblock Conceptnet 5: A large semantic network for relational knowledge.
\newblock In \emph{The People's Web Meets NLP}.

\bibitem[{Tandon et~al.(2018)Tandon, {Dalvi Mishra}, Grus, Yih, Bosselut, and
  Clark}]{propara-emnlp18}
Niket Tandon, Bhavana {Dalvi Mishra}, Joel Grus, Wen-tau Yih, Antoine Bosselut,
  and Peter Clark. 2018.
\newblock Reasoning about actions and state changes by injecting commonsense
  knowledge.
\newblock \emph{EMNLP}.

\bibitem[{Vijayakumar et~al.(2018)Vijayakumar, Cogswell, Selvaraju, Sun, Lee,
  Crandall, and Batra}]{kalyan2016diversekbeam}
Ashwin~K Vijayakumar, Michael Cogswell, Ramprasath~R Selvaraju, Qing Sun,
  Stefan Lee, David Crandall, and Dhruv Batra. 2018.
\newblock Diverse beam search: Decoding diverse solutions from neural sequence
  models.
\newblock \emph{AAAI}.

\bibitem[{Weston et~al.(2016)Weston, Bordes, Chopra, and Mikolov}]{Babi-tasks}
Jason Weston, Antoine Bordes, Sumit Chopra, and Tomas Mikolov. 2016.
\newblock Towards ai-complete question answering: A set of prerequisite toy
  tasks.
\newblock \emph{ICLR}, abs/1502.05698.

\bibitem[{Weston et~al.(2015)Weston, Bordes, Chopra, Rush, van Merri{\"e}nboer,
  Joulin, and Mikolov}]{weston2015towards}
Jason Weston, Antoine Bordes, Sumit Chopra, Alexander~M Rush, Bart van
  Merri{\"e}nboer, Armand Joulin, and Tomas Mikolov. 2015.
\newblock Towards {AI}-complete question answering: A set of prerequisite toy
  tasks.
\newblock \emph{arXiv preprint arXiv:1502.05698}.

\bibitem[{Winograd(1972)}]{winograd72}
Terry Winograd. 1972.
\newblock Understanding natural language.
\newblock \emph{Cognitive Psychology}, 3(1):1--191.

\bibitem[{Yang and Nyberg(2015)}]{Wikihow-sigir2015}
Zi~Yang and Eric Nyberg. 2015.
\newblock Leveraging procedural knowledge for task-oriented search.
\newblock In \emph{SIGIR}.

\end{thebibliography}
\bibliographystyle{acl_natbib}

\end{document}